\documentclass[10pt, final, conference, letterpaper, onecolumn, oneside]{IEEEtran}

\usepackage{cite}
\usepackage{graphicx}
\usepackage{amsmath}
\usepackage{amssymb}
\usepackage{mathtools}
\usepackage{bm}
\usepackage{epsfig}
\usepackage{times}
\usepackage{color}
\usepackage{url}
\usepackage{hyperref}
\usepackage{array}
\usepackage{multirow}
\usepackage{rotating}
\usepackage{extarrows}
\usepackage{mathabx}
\usepackage{wasysym}
\usepackage{ulem}
\usepackage{braket}
\usepackage{pifont}
\usepackage{soul}
\usepackage{enumitem}
\usepackage{float}
\usepackage{tikz}
\usepackage{titlesec}
\usepackage{booktabs}
\usepackage{subcaption}

\usetikzlibrary{positioning}
\titleclass{\subsubsubsection}{straight}[\subsection]
\newcounter{subsubsubsection}[subsubsection]
\renewcommand{\thesubsubsubsection}{\thesubsubsection.\arabic{subsubsubsection}}
\titleformat{\subsubsubsection}
{\normalfont\normalsize\bfseries}{\thesubsubsubsection}{1em}{}
\titlespacing*{\subsubsubsection}
{0pt}{3.25ex plus 1ex minus .2ex}{1.5ex plus .2ex}

\newcolumntype{P}[1]{>{\centering\arraybackslash}p{#1}}
\renewcommand{\arraystretch}{1.5}

\begin{document}
\title{Robust Representation Learning in Masked Autoencoders}%
\author{\IEEEauthorblockN{Anika Shrivastava, Renu Rameshan, and Samar Agnihotri}\\%
Correspondence Email: samar.agnihotri@gmail.com%
}

\maketitle
\begin{abstract}
Masked Autoencoders (MAEs) achieve impressive performance in image classification tasks, yet the internal representations they learn remain less understood. This work started as an attempt to understand the strong downstream classification performance of MAE. In this process we discover that representations learned with the pretraining and fine-tuning, are quite robust---demonstrating a good classification performance in the presence of degradations, such as blur and occlusions. Through layer-wise analysis of token embeddings, we show that pretrained MAE progressively constructs its latent space in a class-aware manner across network depth: embeddings from different classes lie in subspaces that become increasingly separable. We further observe that MAE exhibits early and persistent global attention across encoder layers, in contrast to standard Vision Transformers (ViTs). To quantify feature robustness, we introduce two sensitivity indicators: directional alignment between clean and perturbed embeddings, and head-wise retention of active features under degradations. These studies help establish the robust classification performance of MAEs.
\end{abstract}

\section{Introduction}

Self-supervised learning (SSL) has been shown \cite{data2vec,efficientdet2020} to be a powerful approach for visual representation learning - achieving state-of-the-art performance in various downstream tasks, without relying on costly human annotations. Within SSL, Masked Image Modeling (MIM) \cite{mae2021,convmae2022,beit2022,simmim2022} has proven particularly effective, where the model intentionally masks patches of the input image and then reconstructs the pixels using a sparse set of visible patches, helping models infer global structure of the input from limited context \cite{mae2021}, mirroring the masked-token prediction strategy in NLP \cite{bert}.

Among MIM approaches, the Masked Autoencoder (MAE) \cite{mae2021} is one of the prominent models that brings masked autoencoding to vision---demonstrating strong classification performance. MAE follows an asymmetric encoder–decoder design in which the encoder operates only on the visible patches and lightweight decoder reconstructs the masked content. As the model receives only a small, random subset of patches (often just 25\% of the image), the encoder is forced to learn rich hidden representations. Empirically, the authors of MAE attribute its success to the effectiveness of the pretraining strategy: a Vision Transformer (ViT) \cite{dosovitskiy2020image} trained from scratch requires strong regularizations and long training schedules, whereas MAE pretraining achieves higher classification accuracy (about 84.9\%) with minimal fine-tuning \cite{mae2021}. While pretraining underlies MAE's classification performance, a natural question is how the latent space of pretrained model is organized, particularly with respect to class-wise structure across network depth. Do specific geometric structures arise that aid classification?

While prior work has explored numerous architectural variants and training refinements for MAE, including decoder decomposition \cite{rethinking2024}, mixed input strategies \cite{mixed2022}, and semantic-part learning \cite{semantic2022}--considerably less work has been dedicated to understanding its inner mechanisms \cite{cao2022,kong2023,contrastive1} (described in Section \ref{sec:2.2}). Although, these works provide valuable insights into different aspects of MAE, they do not examine how representations behave internally. In particular, we still lack a clear understanding of how MAE’s encoder organizes class information across network depth, whether any class-specific structure emerges in the pretrained model, despite the absence of labels and how such structure evolves across network depth. While supervised models naturally develop class-separable representations, it is unclear whether similar organization arises in MAE. Moreover, the robustness of the representations obtained after fine tuning, to input perturbations remains underexplored. We believe that MAE’s strong classification performance (compared to ViTs) is fundamentally tied to the structure and behavior of the latent representations formed in the pretrained MAE encoder.

\paragraph{Contributions.} Our main contributions are as follows: 
\begin{enumerate}
    \item We show that pretrained MAE model progressively develops class-separable structure, with distinct class clusters emerging in CLS tokens and also in mean-patch and raw patch token representations across network depth.
    \item We provide a geometric characterization of this behavior through a subspace-based analysis, demonstrating that class-specific subspaces gradually diverge with increasing network depth.
    \item We evaluate the robustness of fine-tuned MAE under Gaussian blur and attention-guided occlusion and demonstrate that classification performance remains stable across a wide range of perturbation levels.
    \item We propose two complementary sensitivity indicators--(i) directional invariance, and (ii) head-wise retention of active features--to quantify the sensitivity of latent representations to perturbations.
    \item Our results reveal that MAE’s classification performance is closely linked to the robustness of its latent representations. We use robustness in the sense that the representation does not change much even when the input is perturbed with controlled blur and occlusion. 
\end{enumerate}

\section{Related Work}

\subsection{\textbf{Masked image modelling}} Masked Image Modeling (MIM) aims to reconstruct missing regions of an image from a corrupted input. Early approaches, such as Stacked Denoising Autoencoders \cite{DAE} treated MIM primarily as a denoising problem. With the advent of Vision Transformer \cite{dosovitskiy2020image}, MIM evolved into a token-prediction paradigm analogous to BERT \cite{bert}, in which an image is partitioned into patches and processed as a sequence. Models like  BEiT \cite{beit2022}, MAE \cite{mae2021}, SimMIM \cite{simmim2022} adopt this formulation. BEiT uses discrete tokens generated by a dVAE tokenizer \cite{dvae}, while MAE and SimMIM demonstrate that reconstruction from raw pixels alone can produce strong representations. Context Autoencoders \cite{CAE} further extend this idea by combining masked representation prediction with masked patch reconstruction in the encoded feature space. Finally, iBOT \cite{ibot} jointly learns a tokenizer via self-distillation over masked patch tokens and class tokens, allowing the model to acquire semantically meaningful representations. A key insight across these works is that vision benefits from higher masking ratios than language: images exhibit strong spatial redundancy, and heavy masking creates a challenging prediction problem that encourages the model to learn global structure rather than rely on local image statistics \cite{mae2021}. 

MAE encodes only visible patches in the encoder, introducing mask tokens only at the decoder stage. This design reduces the computational overhead of the encoder and makes MAE sufficiently scalable. Despite heavy masking during pretraining, MAE learns representations that transfer effectively to classification tasks~\cite{mae2021}, motivating a closer examination of its internal representations.

\subsection{\textbf{Understanding Masked Autoencoders}}
\label{sec:2.2}
Despite MAE's effectiveness, comparatively limited work has examined the structure, evolution, and robustness of the representations learned within its encoder layers. Cao et al. \cite{cao2022} present a theoretical perspective on MAE, showing that its patch-based attention mechanism is equivalent to a learnable integral kernel transform. Kong et al. \cite{kong2023} analyze MAE through a hierarchical latent variable model, demonstrating how masking ratio and patch size influence the semantic level of the learned representations. Other theoretical efforts \cite{contrastive1,contrastive2,contrastive3} relate MAE to contrastive learning. In particular,\cite{contrastive1} shows that MAE implicitly aligns mask-induced positive pairs and introduces a uniformity-enhanced loss to address dimensional collapse. Both \cite{contrastive2} and \cite{contrastive3} present interpretive frameworks that highlight MAE’s invariance to random masking-- the work in \cite{contrastive2} does so by reformulating masked image modeling as an equivalent Siamese framework, interpreting MAE as a special case of contrastive learning that reveals occlusion-invariant features, while \cite{contrastive3} employs a local contrastive framework to analyze both the reconstructive and contrastive aspects of MAE.

Most work on MAE focuses on modifying its architecture or training strategy to improve performance. The work \cite{rethinking2024} replaces self-attention layers in the decoder with cross-attention to aggregate encoder output into each input token within the decoder layers, while \cite{mixed2022} replaces masked tokens with visible tokens from another image to reconstruct two original images from a single mixed input; the work \cite{semantic2022} incorporates semantic-part supervision into MAE training; and \cite{convmae2022} integrates convolutional modules into a multi-scale hierarchical design.

While these works address different aspects of MAE and introduce improvements to its architecture or training, they do not explicitly examine the learned representations that ultimately drive its strong classification performance. Moreover, the behaviour of MAE's representations under input perturbations remains relatively less explored. This gap motivates our work. We show that the pretrained MAE encoder develops a class-discriminative structure, with representations from different classes becoming increasingly separable in the final encoder layers. This structure provides a strong initialization for fine-tuning and leads to more robust and stable representations for classification.

\section{Proposed Analysis Pipeline}

This section describes our analysis pipeline for studying the internal representations learned by MAE through pretraining and fine-tuning. We first analyze the layer-wise token embeddings in the pretrained MAE encoder to examine how class-level structure evolves across network depth, with the aim of assessing whether class-relevant organization emerges in the absence of supervision. We then extend this analysis to the fine-tuned model and study the impact of input perturbations on both classification performance and the robustness of latent representations. Representation robustness is characterized using two complementary indicators: directional alignment between clean and perturbed embeddings, and feature-level robustness within individual attention heads.

\subsection{Layer-wise structural analysis of the pretrained encoder}
\label{sec:structural}

\subsubsection{Overview of MAE:}
Given an input image \(x \in \mathbb{R}^{H \times W \times 3}\), MAE first divides it into \(N\) non-overlapping patches of size \(n \times n\), forming a sequence \(x_p = \{x_p^1, x_p^2, \ldots, x_p^N\},\) where each \(x_p^i \in \mathbb{R}^{n \times n \times 3}\). A random subset of \(N_v\) visible patches \(x_v \subset x_p\) is selected based on the masking ratio (typically $75$\%). Each patch is flattened and projected into a \(D\)-dimensional embedding by the embedding layer \(E_p : \mathbb{R}^{n \times n \times 3} \rightarrow \mathbb{R}^D\). A learnable class (CLS) token \(x_{cls} \in \mathbb{R}^{D}\) is prefixed, and positional embeddings are added to preserve spatial information. The resulting sequence is passed through the encoder \(f(.)\), producing the latent representation \(z = f([x_{cls}; E_p(x_v)] + E_{pos})\), where \(E_{pos} : \mathbb{R}^{(N_v + 1) \times D}\). These encoded tokens are concatenated with a set of learned mask tokens \(e_M\) and fed into a lightweight decoder \(g(.)\), which reconstructs the masked patches to yield \(\hat{x} = g(z, e_M)\). We use the MAE ViT-Base architecture \cite{mae2021}, consisting of $12$ transformer encoder layers with an overall embedding dimension of \(D = 768\). Every layer contains $12$ self-attention heads with embedding dimension $d_h = 64$. The output from all heads are concatenated to form the full $768$-dimensional embedding at each layer. We follow the original MAE setup, using an input resolution of $224 \times 224$ and a masking ratio $0.75$, resulting in $N_v = 49$ visible patches.

\subsubsection{Layer-wise embedding extraction:}
To study how MAE organizes class information across network depth, we extract token embeddings from each encoder layer \(l \in \{1, \ldots, 12\}\).
For an input image \(x\), the layer-wise output of every encoder layer \(l\) is:
\[
z^{(l)} = \{ z_0^{(l)}, z_1^{(l)}, \ldots, z_{N_v}^{(l)} \} \in \mathbb{R}^{(N_v + 1) \times D},
\]
where $z_0^{(l)}$ denotes the CLS embedding and $z_i^{(l)}$ for $i \geq 1$ correspond to the visible patch embeddings. Using this, we examine three types of embeddings and track how they evolve with network depth: CLS embedding $z_{\mathrm{0}}^{(l)} \in \mathbb{R}^D$, patch token embeddings $\{z_i^{(l)}\}_{i=1}^{N_v}$ and their mean patch embedding
\(\bar{z}^{(l)} = \frac{1}{N_v} \sum_{i=1}^{N_v} z_i^{(l)}\) which aggregates information across all visible patches into a single global descriptor of the image and is known to serve as an effective representation in classification tasks \cite{mae2021}. Motivated by this property, we adopt it in our fine-tuned MAE analysis when evaluating robustness under perturbations.

In Section \ref{sec:layerwise}, we use t-SNE \cite{tsne} to visualize the evolution of token embeddings as the input propagates through successive self-attention layers. While these visualizations provide qualitative insight into the emergence of class-related structure, they do not yield a geometry-aware characterization of how representations are organized across classes and network depth. We therefore complement this analysis with a subspace-based geometric study.

\subsubsection{Subspace geometry:}
To study the class-level geometric structure of MAE representations, we collect the patch embeddings from all images belonging to the same class and arrange them into a matrix. For class $\mathcal{C}$, the layer-$l$ representation matrix is defined as $ X_{\mathcal{C}}^{(l)} \in \mathbb{R}^{N_c \times D}$, 
where each row is a patch token embedding of dimension $D$ and $N_c$ is the total number of visible patch tokens aggregated across all images of class $\mathcal{C}$. 

To extract the dominant geometric structure of these representations, we apply singular value decomposition (SVD) to $X_{\mathcal{C}}^{(l)}$, yielding 
\( X_{\mathcal{C}}^{(l)} = U_{\mathcal{C}}^{(l)} \, \Sigma_{\mathcal{C}}^{(l)} \, {V_{\mathcal{C}}^{(l)}}^{\top}\).
The right singular vectors \({V_{\mathcal{C}}^{(l)}}^{\top}\) define orthogonal directions in feature space along which the embeddings of class $\mathcal{C}$ exhibit the largest variance. Retaining the top-$k$ singular vectors gives the best possible basis for representation without redundancy yielding a compact, low-dimensional subspace $S_{\mathcal{C}}^{(l)} = \text{span}\big(V_{{\mathcal{C}},k}^{(l)}\big)$ for class \(\mathcal{C}\) at depth $l$. 

In Section \ref{sec:layerwise}, we examine how these class-specific subspaces are oriented relative to one another in the feature space using prinicpal angles \cite{zhu2012angles}. By tracking the principal angles between the subspaces across layers, we obtain a layer-wise view of how class-specific embeddings evolve and are separated with increasing depth in the pretrained MAE encoder. 

\subsection{Fine-tuning robustness}

\subsubsection{Perturbations.}
We analyze the classification performance of a fine-tuned MAE under two types of perturbations, each applied at multiple severity levels:
\begin{enumerate}
    \item \textbf{Gaussian blur}: Blur severity is increased by using larger kernel sizes and standard deviations, which lead to a monotonic decrease in PSNR and SSIM. This allows blur severity to be ordered by the resulting image degradation, rather than by parameter choice alone.
    \item \textbf{Attention-guided occlusion}: Instead of masking random regions, we use attention rollout \cite{abnar2020quantifying} to estimate the contribution of each patch to the model’s final prediction. Attention rollout recursively multiplies attention matrices across layers to produce a global importance score for each patch. Patches are then ranked according to these rollout scores, and the top $p\%$ most attended patches--corresponding to different masking levels--are masked to obtain the occluded image $x_o$. This perturbation explicitly targets regions that the MAE attends to most, providing a challenging setting for assessing the model robustness.
\end{enumerate}

\subsubsection{Representational robustness indicators:}
\label{sec:stability}

To study how MAE’s latent representations respond to input perturbations, we examine their robustness from two complementary perspectives:

\paragraph{Directional Robustness:}

As an initial analysis, we examine how MAE’s latent representations behave under different random masked versions of the same image. For a fixed image, we generate $100$ randomly masked images at the same masking ratio. Across these runs, the resulting latent embeddings remain highly aligned in direction. This behavior is consistently observed across multiple images: while both direction and magnitude vary from image-to-image, they remain remarkably consistent across masked variants of the same image. In particular, the magnitudes for a given image remain tightly concentrated (e.g., in the range $92.4$ to $95.4$ for one example) across all its masked variants. These findings suggests that learned latent representations are largely invariant to random masking, consistent with the findings in \cite{contrastive2,contrastive3}. Building on this observation, we next examine whether similar robustness extends to perturbations such as blur and occlusion, where the severity of the perturbation is systematically increased.

We use cosine similarity measure for assessing robustness in terms of direction.
\noindent Formally, let $\bar{z}_{(clean)} \in \mathbb{R}^D$ denote the mean patch embedding for the clean input, and let $\bar{z}_{(pert)} \in \mathbb{R}^D$ be the corresponding embedding under blur or occlusion. Directional alignment is quantified as: 
\begin{equation}
    \cos(\theta) = 
    \frac{\langle \bar{z}_{(clean)}, \bar{z}_{(pert)} \rangle}
    {\| \bar{z}_{(clean)} \|_2 \, \| \bar{z}_{(pert)} \|_2}.
\end{equation}
A value closer to $1$ indicates that the encoder projects degraded inputs in a direction similar to that of the clean input in latent space, reflecting strong directional consistency.

\paragraph{Feature-level robustness of attention-heads:}
\noindent
While cosine similarity captures directional alignment between clean and perturbed embeddings, it does not capture feature-level changes within individual attention heads. To address this limitation, we analyze the set of active features (defined below) retained within each attention head under input perturbations.
For a given layer $l$ and attention head $h$, the head-wise output is given by:
\begin{equation}
    O^{(l,h)} = A^{(l,h)} V^{(l,h)}, \quad O^{(l,h)} \in \mathbb{R}^{N_v \times d_h},
\end{equation}
where $A^{(l,h)} \in \mathbb{R}^{N_v \times N_v}$ is the attention matrix and $V^{(l,h)} \in \mathbb{R}^{N_v \times d_h}$ is the value matrix, with $d_h = 64$ for ViT-Base. 
Each row of $V^{(l,h)}$ corresponds to the value vector of a visible patch token, i.e., a $d_h$-dimensional representation, where each dimension corresponds to a feature. Thus, in $V^{(l,h)}$, each patch is represented by a $64-D$ feature vector, and each column of V represents the activation of a particular feature across all visible tokens.
The attention matrix $A^{(l,h)}$ determines which patches contribute in updating a given patch's representation. Specifically, the $i^{\text{th}}$ row of $A^{(l,h)}$ is a probability distribution over visible patch tokens, indicating how strongly each patch contributes to the updated representation of patch $i$. The head output for token $i$
\begin{equation}
    O_{i,k}^{(l,h)} = \sum_{j=1}^{N_v} A_{i,j}^{(l,h)} V_{j,k}^{(l,h)}, \qquad k = 1,\dots,d_h.
\end{equation}
The resulting vector $O_{i,:}^{(l,h)} \in \mathbb{R}^{1 \times d_h}$ represents the updated $d_h$-dimensional representation of the $i^{\text{th}}$ patch, obtained as a weighted aggregation of features from the attended patches.
In this sense, the head output reflects a combination of importance (via attention) and feature presence (via the value matrix). A feature becomes prominent in $O_{i,:}^{(l,h)}$ only if it has high magnitude in the value representations of patches that receive high attention. Conversely, if a feature is absent across the attended patches, its magnitude in the updated representation remains low or zero. We refer to the features with high magnitude in $O_{i,:}^{(l,h)}$ as \textit{active} features. The outputs of all $12$ heads act as a fundamental building blocks as they are concatenated at each encoder layer to form a $768$-dimensional layer embedding, that is passed to subsequent layers, ultimately yielding the final representation used for classification. 

We take the mean over all rows of $O^{(l,h)}$, i.e., calculating mean patch token $\bar{z}$ to obtain a single 64-dimensional vector per attention head for each image. For a given class $\mathcal{C}$, we collect these vectors across all images in the class and, for each image we identify the top-$k$ active features based on magnitude. A feature is considered \textit{common} within class $\mathcal{C}$ if it appears among the top-$k$ features for at least $60\%$ of the images. For example, for a class containing $50$ images and $k=10$, the common features are those dimensions of $\bar{z}$ that are present among the top-$10$ active features in at least $30$ images. The number of such features gives the common-feature count \(C_{clean}^{(l,h)}\) which quantifies how many features remain consistently active within layer $l$, head $h$ across images of class $\mathcal{C}$. 
We repeat the same procedure for perturbed inputs to obtain \(C_{pert}^{(l,h)}\). Importantly, we keep these counts as the cardinality of the intersection between the clean head-level features and the features activated under perturbation. This directly quantifies how reliably each head preserves its clean feature activations. To make the comparison visually clearer, we also compute the mean drop in feature count:
\(
\Delta C^{(l,h)} = C_{\text{clean}}^{(l,h)} - C_{\text{pert}}^{(l,h)}
\) averaged over all layers and heads for a given  perturbation level.

In Section \ref{sec:similarity_results}, we present the empirical findings obtained using the similarity indices define above.

\section{Experiments and Results}
\label{sec:results}

In this section, we present both qualitative and quantitative analyses to trace how MAE’s semantic representations evolve across network depth in a pretrained encoder and how they behave under perturbations after fine-tuning. All experiments are conducted on the Imagenet-1K dataset \cite{imagenet}, using the MAE ViT-B model \cite{mae2021} with default architectural settings. Wherever a different dataset is used to support or verify a specific result, it is explicitly mentioned at that point.

\subsection{Layer-wise analysis of representational structure of MAE pre-trained encoder}
\label{sec:layerwise}

\subsubsection{Evolution of token embeddings across depth:}
Following Section \ref{sec:structural}, we consider a fixed set of ten ImageNet-1K classes \cite{imagenet}, selected to include both semantically similar categories (e.g., multiple dog breeds) and diverse categories (e.g., balloon, mosque, speedboat), and draw a total of 100 images uniformly at random (10 per class). We then extract the CLS embeddings $z_{\mathrm{0}}^{(l)}$, raw patch token embeddings $\{z_i^{(l)}\}_{i=1}^{N_v}$, and their mean patch embeddings \(\bar{z}^{(l)}\) for each layer $l$, and visualize them using t-SNE \cite{tsne}. 

\begin{figure}[t!]
    \centering
    \begin{subfigure}[t]{0.47\textwidth}
      \centering
      \includegraphics[width=\textwidth]{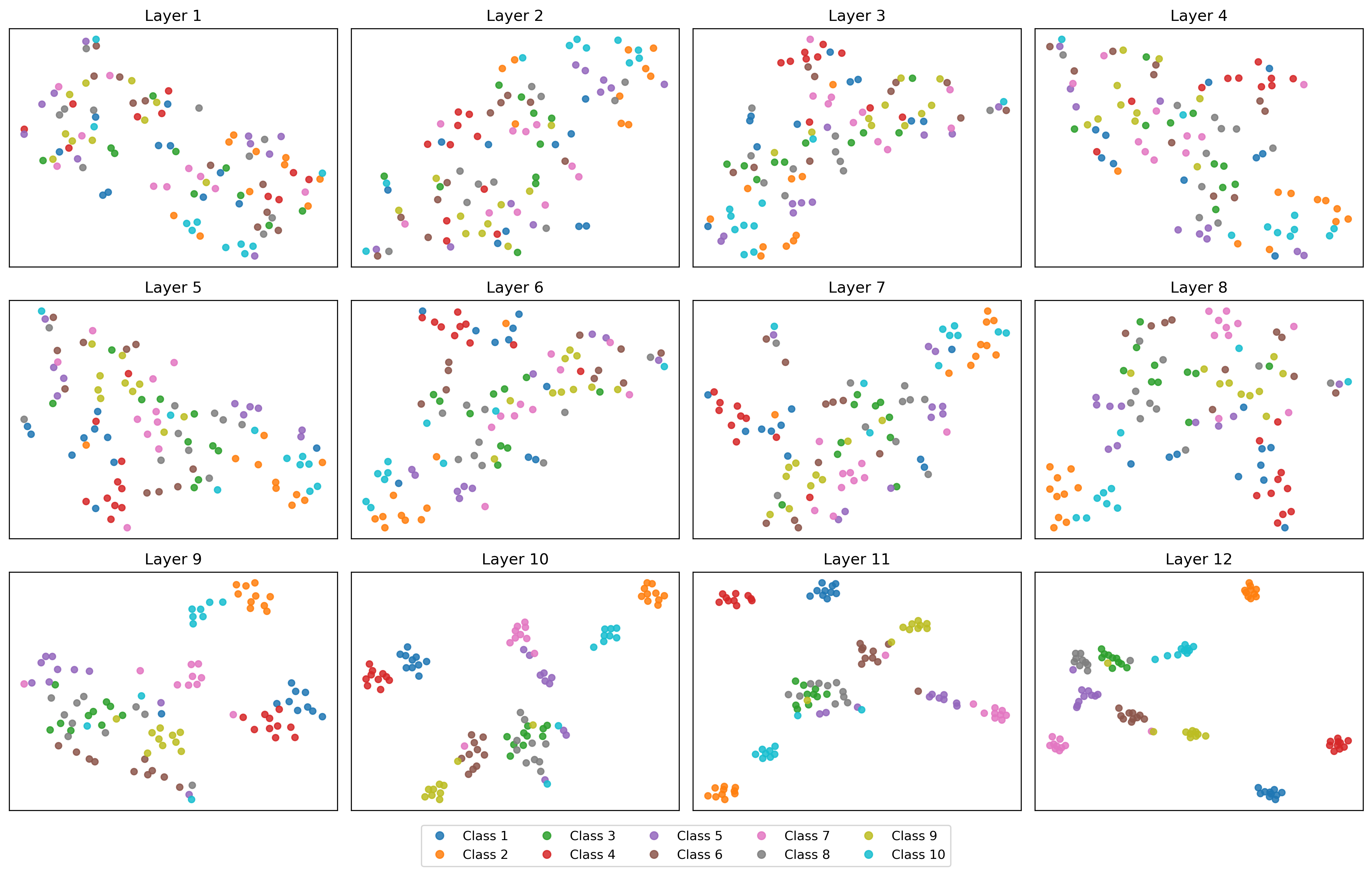}
      \caption{CLS tokens.}
      \label{fig:clstoken}
    \end{subfigure}
    \hfill
    \begin{subfigure}[t]{0.47\textwidth}
      \centering
      \includegraphics[width=\textwidth]{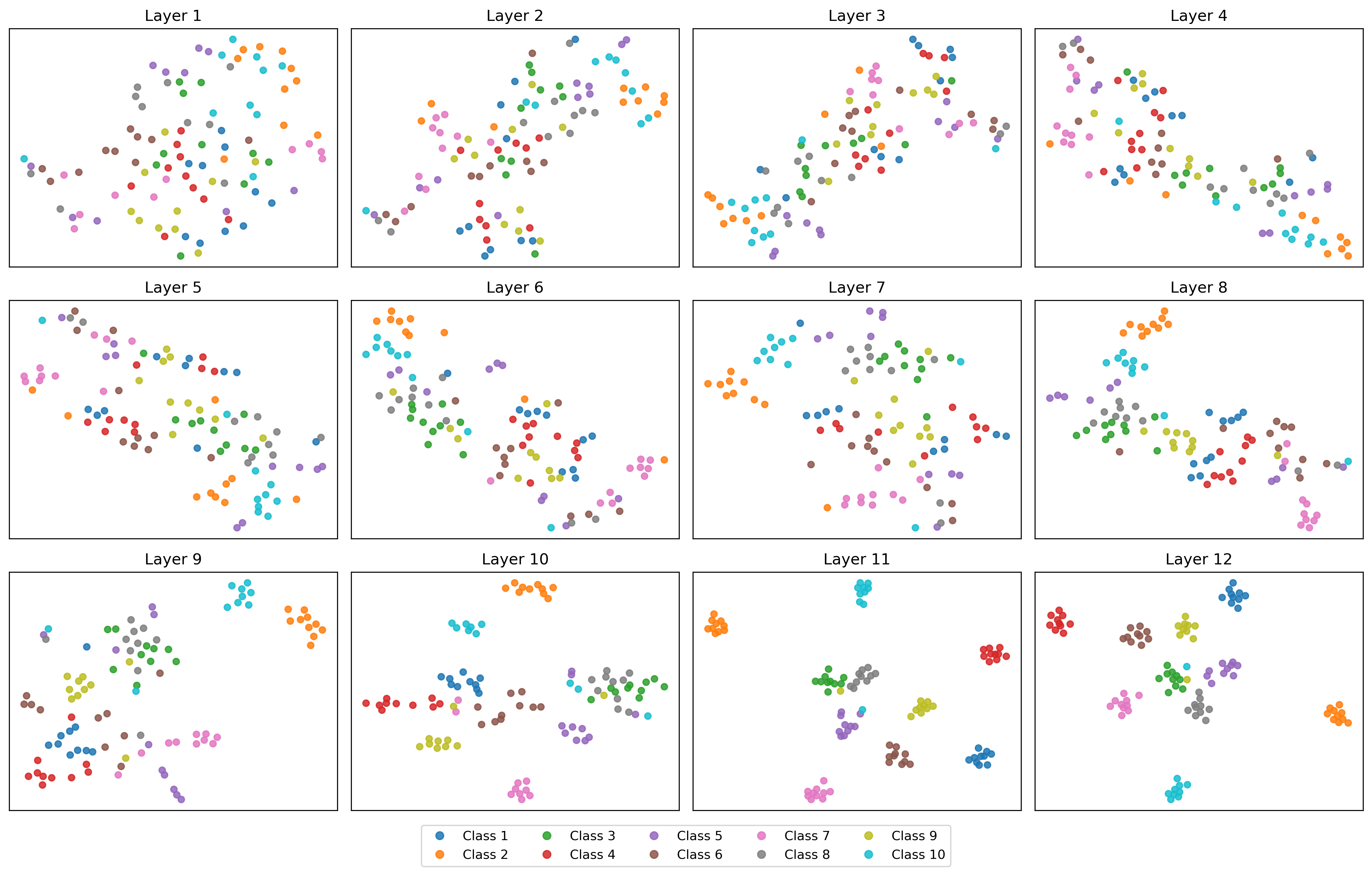}
      \caption{Mean patch tokens.}
      \label{fig:meantoken}
    \end{subfigure}
    \caption{t-SNE visualizations of token embeddings across encoder layers: 
    (a) CLS tokens and (b) mean patch tokens.}
    \label{fig:tsne_combined}
\end{figure}

In the early encoder layers, embeddings from different classes occupy largely overlapping regions with no clear class separation. However, around layers \(l = 9-10\), distinct class clusters begin to emerge, becoming progressively more separable in the deeper layers. The same trend is consistently observed for the CLS token (Fig. \ref{fig:clstoken}), the mean patch token (Fig. \ref{fig:meantoken}), and, perhaps most interestingly for the raw patch tokens as well. The fact that even individual patches exhibit class-level separation, indicates that, through self-attention, patch representations become increasingly contextualized, allowing class-discriminative information to be distributed across tokens.

\begin{figure}[t!]
    \centering
    \includegraphics[width=0.95\textwidth]{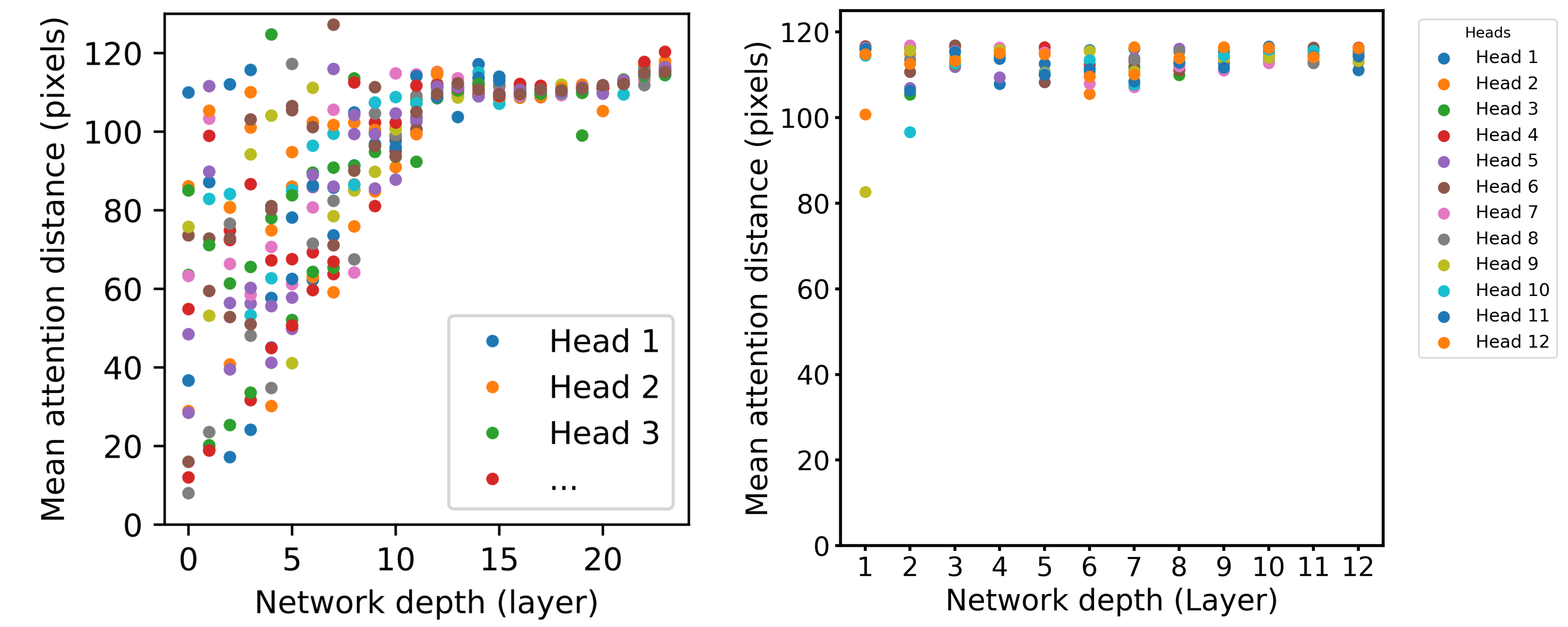}
    \caption{Each dot shows the mean attention
    distance across images for one head in \textbf{left:} ViT (adapted from \cite{dosovitskiy2020image}) and \textbf{right:} MAE.}
    \label{fig:attndist}
\end{figure}

We hypothesize that the high masking ratio used during MAE pretraining contributes to the clustering behaviour. With only a small fraction of patches visible, the encoder cannot rely on local neighborhoods and is forced to attend over far away patches from early layers. To quantify this behavior, we compute mean attention distances \cite{heo2021rethinking} across $100$ images from ten ImageNet-1K classes, for each attention head across the $12$ layers. In standard ViTs, attention expands with network depth: initial layers focus on local patterns, while deeper layers gradually incorporate more global context (Fig.~\ref{fig:attndist} (left)). In contrast, for MAE (even with no masking), we find consistently high mean attention distances of approximately $80$–$120$ pixels across all heads and layers (Fig.~\ref{fig:attndist} (right)), indicating that MAE attends globally from the outset. A similar attention-distance pattern is observed on $100$ randomly selected images from the SAM~\cite{sam} dataset, with values exceeding $110$-$120$ pixels across all heads and layers. This persistent long-range attention offers a mechanistic explanation for the emergence of class structure observed in the t-SNE plots.

\subsubsection{Class-wise subspace geometry across network depth:}
To complement the qualitative observations from t-SNE, we adopt a geometric viewpoint to quantify how class-specific structure emerges in the latent space of a pretrained MAE. The core idea is to treat the collection of patch-token embeddings from a class as points in $\mathbb{R}^{768}$ and study the subspaces these points span.

\begin{figure}[t!]
    \centering
    \includegraphics[width=1.0\textwidth]{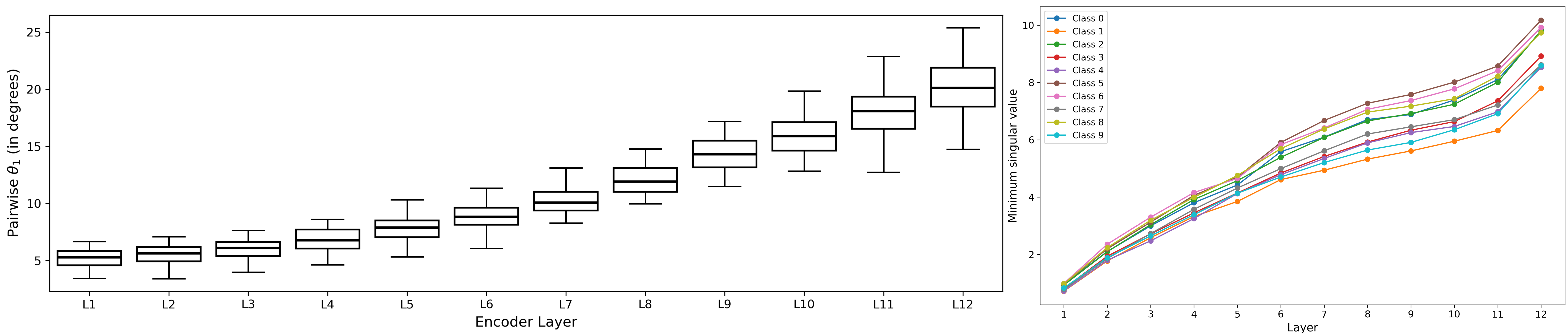}
    \caption{\textit{left:} Layer-wise distribution of principal angles $\theta_1$ (in degrees) between classes across layers. \textit{right:} Layer-wise evolution of the minimum singular value across classes.}
    \label{fig:principal_angles}
\end{figure}

We use the selected ImageNet-1K classes $\mathcal{C}_i$, $i \in \{1, 2, \dots, 10\}$ for this analysis and randomly sample $50$ images per class. As described in Section \ref{sec:structural}, at each encoder layer $l$, visible patch embeddings from class $\mathcal{C}_i$ are stacked and SVD is applied to yield a $k$-dimensional class-specific subspace by retaining the top-$k$ dominant directions of variance. Restricting the analysis to the leading $k$ directions is more for covenience of analysis. We compute principal angles between every pair $(S_{\mathcal{C}_i}^{(l)}, S_{\mathcal{C}_j}^{(l)})$. The smallest principal angle $\theta_1$ serves as a measure of subspace proximity: small angles indicate stronger alignment (i.e., overlapping subspaces), while larger angles reflect greater separation. Figure~\ref{fig:principal_angles} (\textit{left}) shows the distribution of $\theta_1$ over all class pairs, across layers using box plots. In the early layers, the distributions are tightly concentrated at low angles, indicating that subspaces corresponding to different classes remain closely aligned in feature space. With increasing depth, the distribution of $\theta_1$ shifts progressively toward larger values, with both the median and the interquartile range increasing across layers. This indicates that class-specific subspaces systematically rotate away from one another and become increasingly well separated in deeper layers. \textit{We further observe that singular values also increase with depth, including the minimum singular value} (Figure 3 (\textit{right})), \textit{suggesting that the encoder progressively selects more significant basis directions that contribute meaningfully to the class-specific structure in deeper layers.} 

Taken together, we show a clear layer-by-layer emergence of class-specific structure: as the network depth increases, embeddings from different classes gradually diverge and occupy increasingly distinct subspaces in the latent space.

\subsection{Robustness of fine-tuned MAE under input perturbations}
\label{sec:robustness}

Having established that the pretrained MAE encoder develops class-level structure, we now examine the behavior of these representations after supervised fine-tuning. Our goal is to assess the robustness of model's classification performance under controlled input degradations. We compute mean classification accuracy over the selected ImageNet-1K classes ($50$ images per class) using predictions based on the mean patch embedding $\bar{z}^{(l)}$. We also evaluate the model on Caltech-256~\cite{caltech} dataset, achieving a top-$1$ accuracy of $89.47\%$ on clean inputs.

\begin{table}[t!]
    \caption{Top-$1$ accuracy (mean $\pm$ $\sigma$), PSNR, and SSIM for varying blur levels, characterized by kernel size and standard deviation on ImageNet-1K dataset.}
    \label{tab:blur_metrics}
    \centering
    \scriptsize
    \renewcommand{\arraystretch}{1.2}
    \setlength{\tabcolsep}{10pt}
    \begin{tabular}{|c|c|c|c|c|}
    \hline
    \textbf{Blur level} & \textbf{Blur setting ($k$,$s$)} & \textbf{Top-1 accuracy (\%)} & \textbf{PSNR (dB)} & \textbf{SSIM} \\
    \hline
    I   & $k=5,\, s=1.0$   & $89.790 \pm 0.23$ & 28.12 & 0.868 \\
    II  & $k=5,\, s=2.0$   & $88.000 \pm 0.20$ & 25.62 & 0.774 \\
    III & $k=5,\, s=4.0$   & $87.500 \pm 0.48$ & 25.01 & 0.743 \\
    IV  & $k=5,\, s=9.0$   & $87.400 \pm 0.62$ & 24.85 & 0.734 \\
    V   & $k=7,\, s=2.0$   & $82.600 \pm 0.72$ & 24.69 & 0.729 \\
    VI  & $k=7,\, s=4.0$   & $82.400 \pm 0.39$ & 24.21 & 0.667 \\
    VII & $k=7,\, s=13.5$  & $82.300 \pm 0.34$ & 23.65 & 0.646 \\
    VIII& $k=7,\, s=15.0$  & $81.202 \pm 0.55$ & 23.00 & 0.643 \\
    IX  & $k=11,\, s=2.0$  & $80.800 \pm 0.46$ & 21.40 & 0.601 \\
    X   & $k=11,\, s=5.0$  & $80.800 \pm 0.77$ & 20.06 & 0.466 \\
    \hline
    \end{tabular}
\end{table} 

\begin{figure}[t!]
    \centering
    \includegraphics[width=0.75\textwidth]{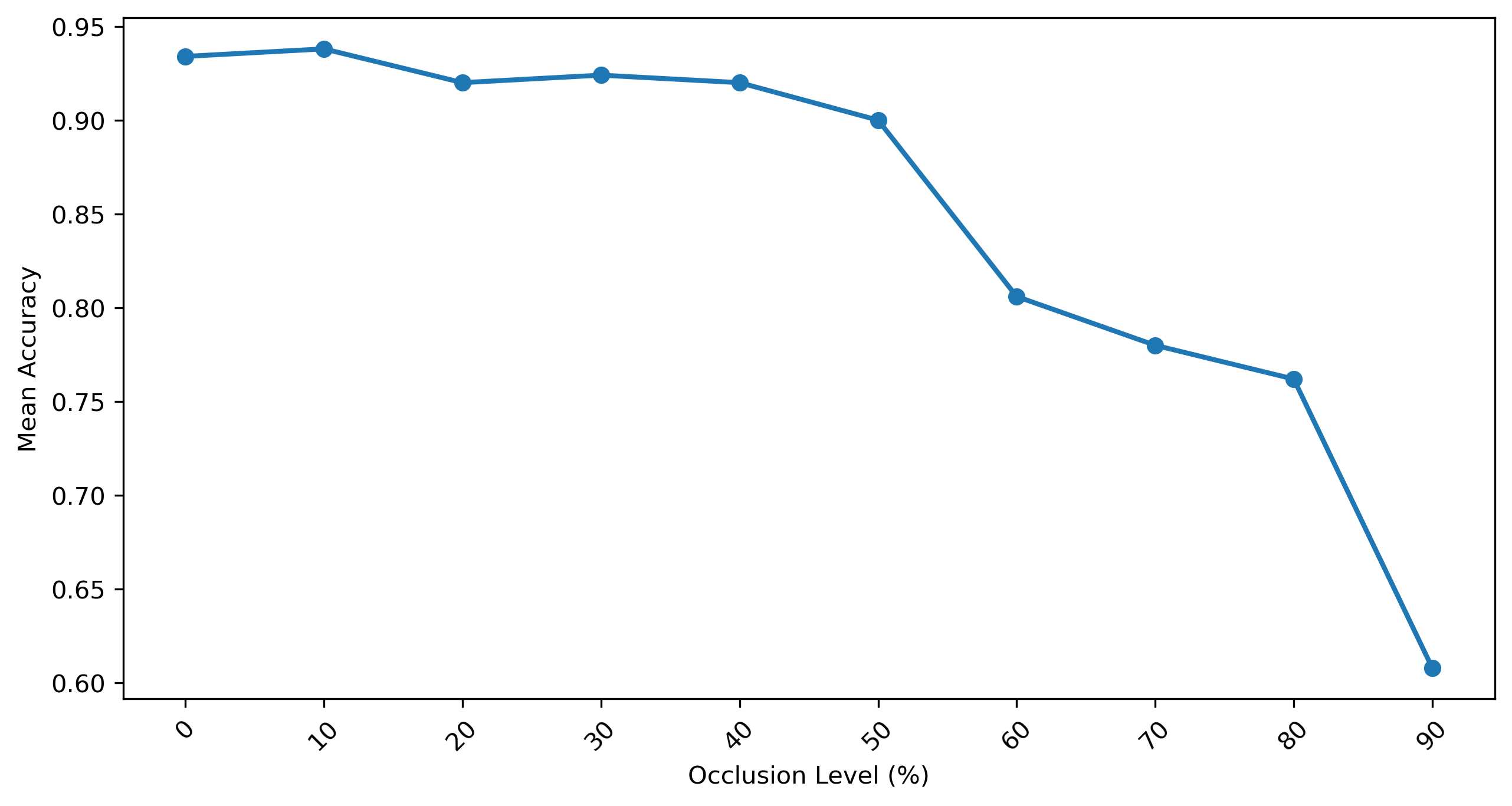}
    \caption{Occlusion level vs Mean Accuracy plot on ImageNet-1K dataset.}
    \label{fig:occlusion}
\end{figure}

Table \ref{tab:blur_metrics} reports the PSNR, SSIM, and top-$1$ accuracy (mean $\pm$ standard deviation ($\sigma$) across multiple runs) under Gaussian blur for different blur levels. Although image quality drops sharply—PSNR falling from $28.12$ dB to $20.06$ dB and SSIM from $0.868$ to $0.466$—the top-$1$ accuracy remains comparatively stable, remaining above $80$\% even for the strongest blur. This empirically indicates that even when local details are heavily degraded, the latent representation still retains the information necessary for correct classification.

To probe robustness under more extensive information loss, we use attention-guided occlusion, with results averaged over multiple runs and exhibiting low variability. Despite this adversarial masking strategy, fine-tuned MAE shows high accuracy even when $50$\% of the most attended patches are removed, as can be seen from Figure~\ref{fig:occlusion}. Beyond $60$\% occlusion, accuracy drops sharply but remains at $60.8$\% even when $90$\% of the most attended patches are masked. This behaviour indicates that MAE does not rely solely on the most attended patches for classification. Instead, the model is able to form a semantic representation from whichever subset of patches remains visible.
This observation is consistent with our pretraining analysis, where raw patch tokens exhibited clear class-level separation, showing that patch embeddings encode class-relevant information. 

We further compute results on the ImageNet-C~\cite{imagenet-c} dataset which includes $15$ degradation types spanning noise, blur, weather, and digital artifacts, each evaluated at five severity levels to assess MAE's robustness under a broader set of degradations. MAE maintains stable classification performance (above 75\%) for many degradations, including all weather-based variants, defocus blur, motion blur, contrast, pixelation and JPEG compression, with a gradual drop as severity increases. Noise-based degradations lead to a steeper yet progressive decline in accuracy. Overall, these results indicate that MAE exhibits strong robustness to a wide range of algorithmically generated degradations. In contrast, MAE struggles on datasets involving larger distribution shifts, including ImageNet-R~\cite{imagenet-r} which alters texture and local image statistics through artistic renditions, and ImageNet-A~\cite{imagenet-a} which consists of adversarial natural images drawn from a shifted input distribution.

\subsection{Robustness of latent representations}
\label{sec:similarity_results}

\begin{table}[t!]
    \centering
    \caption{Mean cosine similarity between clean and perturbed embeddings for (a) Gaussian blur and (b) attention-guided occlusion on ImageNet-1K.}
    \label{tab:cosine_combined}
    \vspace{0.5em}
    \renewcommand{\arraystretch}{1.15}
    \setlength{\tabcolsep}{3pt}  
    
    \textbf{(a) Gaussian blur} \\[4pt]
    \resizebox{\textwidth}{!}{
    \begin{tabular}{|c|c|c|c|c|c|c|c|c|c|c|}
    \hline
    \textbf{Level} & I & II & III & IV & V & VI & VII & VIII & IX & X \\
    \hline
    \textbf{Mean sim.} 
    & 0.966 & 0.911 & 0.908 & 0.907 & 0.884 & 0.866 & 0.859 & 0.852 & 0.793 & 0.790 \\
    \hline
    \end{tabular}
    }
    
    \vspace{1.5em}
    
    \textbf{(b) Attention-guided occlusion} \\[4pt]
    \resizebox{\textwidth}{!}{
    \begin{tabular}{|c|c|c|c|c|c|c|c|c|c|c|}
    \hline
    \textbf{Occlusion (\%)} & 0 & 10 & 20 & 30 & 40 & 50 & 60 & 70 & 80 & 90 \\
    \hline
    \textbf{Mean sim.} & 1.000 & 0.954 & 0.856 & 0.855 & 0.855 & 0.852 & 0.836 & 0.805 & 0.790 & 0.665 \\
    \hline
    \end{tabular}
    }
    
    \setlength{\tabcolsep}{6pt}
\end{table}
    
We now report the empirical observations based on the robustness indicators defined in Section~\ref{sec:stability}. We first examine how the direction of latent embeddings changes under perturbations, by taking mean cosine similarity over the selected ImageNet-1K classes $\mathcal{C}_i$ to measure directional alignment of $\bar{z}_{pert}$ with respect to $\bar{z}_{clean}$. Despite increasing degradation levels, the similarity remains high under moderate blur and occlusion. Table~\ref{tab:cosine_combined}(a) shows cosine similarity stays above $0.85$ for all blur levels up to $(k=7,\, \sigma=15.0)$; at the strongest blur setting ($k=11, \sigma=5.0$), it drops only to $0.79$, consistent with the modest decrease in accuracy (Table~\ref{tab:blur_metrics}). A similar pattern appears under occlusion (Table~\ref{tab:cosine_combined}(b)), where similarity remains relatively high upto $60\%$ occlusion and then declines gradually, consistent with the drop observed in accuracy curve (Figure~\ref{fig:occlusion}). 

These results show that under moderate perturbations, MAE continues to project degraded inputs to latent embeddings whose directions remain closely aligned with their clean counterparts. However, as severity increases, this alignment progressively drops from above $0.85$ to $0.790$ for blur and $0.665$ for occlusion at highest perturbation setting. Experimentally, we also observe that misclassified images exhibit a much lower cosine similarity with their clean embeddings as compared to the correctly classified ones: For highest blur level, correctly classified cosine is $\approx 0.915$ whereas misclassified cosine $\approx 0.289$; a similar trend is observed for occlusion as well. These observations suggest that directional alignment in the latent space is closely linked to the classification accuracy of MAE, supporting loss of robustness at higher perturbations, observed in Section \ref{sec:robustness}. In addition to directional robustness, we also find that across most blur and occlusion levels, the absolute difference between the norm of the perturbed embedding and that of the clean embedding remains relatively low (e.g., the absolute difference $\approx 1.23$–$5.56$, for one example), indicating the magnitudes of embeddings remain tightly concentrated under perturbations as well.

For examining robustness at a finer scale, we analyze whether features that are consistently active for clean inputs within each attention head continue to remain active when the input is perturbed. For each layer–head pair, we compare the common-feature counts obtained from clean input with those under blur or occlusion. These common features are identified based on their consistent activation across clean images of a class. By checking whether the same features remain active after perturbation, we directly assess how much of the head’s original feature activations is preserved when the input is degraded. For clean inputs, shallow layers exhibit higher common-feature counts than deeper layers, and this behavior is largely preserved under perturbations as well. 

Under blur, feature retention remains largely robust across layers and heads. For most heads, \(C_{pert}^{(l,h)}\) stays close to the clean baseline, with noticeable drops appearing only under the strongest blur. This indicates that moderate blur does not meaningfully disrupt the consistent feature activation structure observed for clean inputs. In contrast, occlusion produces a much sharper effect. Up to $50$--$60\%$ masking, a large portion of clean features remain active. Beyond this point however, feature retention drops rapidly, especially in deeper layers, where several heads lose most of their common active features. This collapse mirrors the steep accuracy drop observed in Figure~\ref{fig:occlusion}.

\begin{figure}[t!]
    \centering
    \begin{subfigure}{0.48\textwidth}
      \centering
      \includegraphics[width=\textwidth]{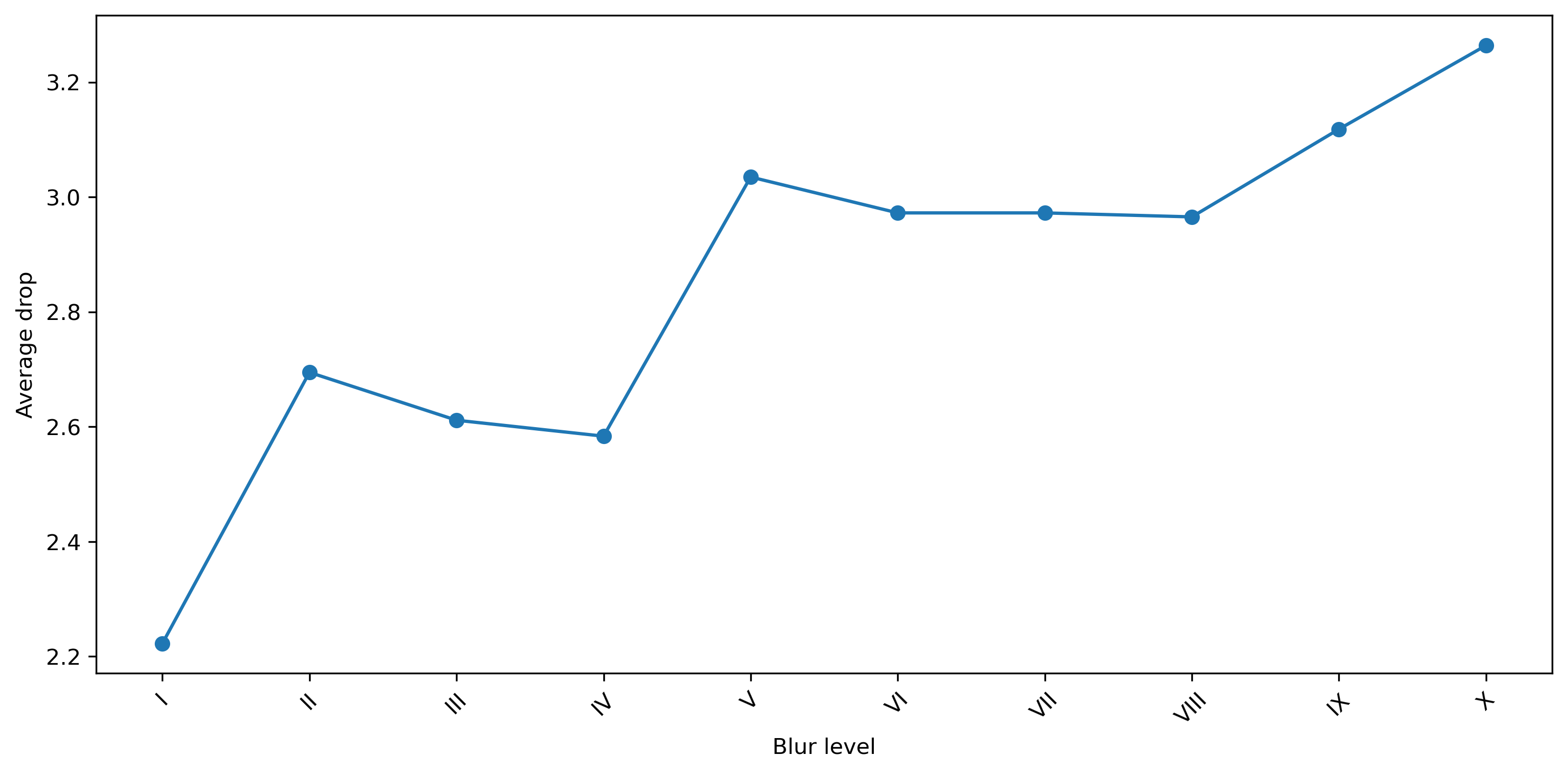}
      \caption{Blur}
      \label{fig:blur_drops}
    \end{subfigure}
    \hfill
    \begin{subfigure}{0.48\textwidth}
      \centering
      \includegraphics[width=\textwidth]{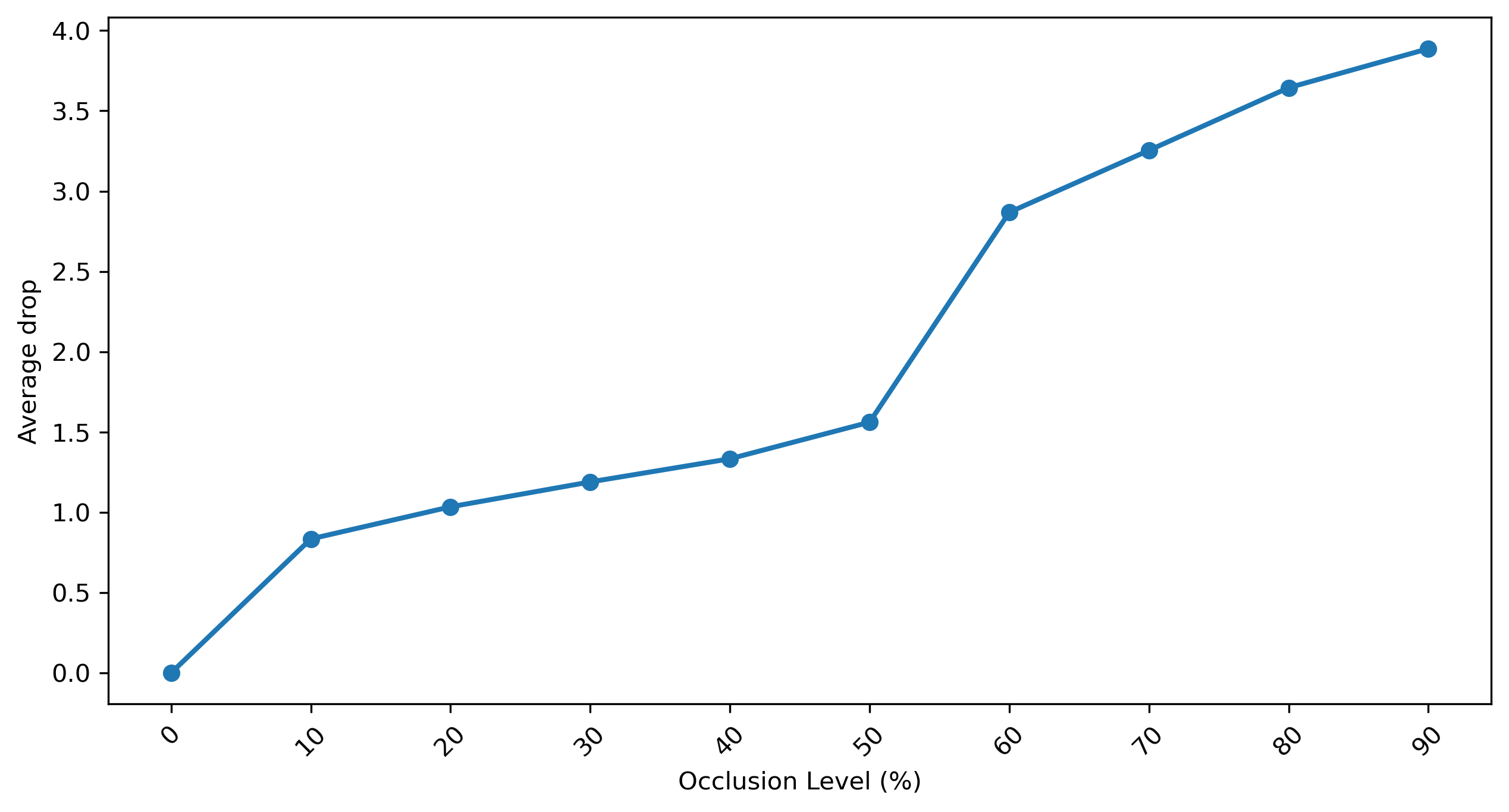}
      \caption{Occlusion}
      \label{fig:occlusion_drops}
    \end{subfigure}
    \caption{Average drop from $C_{\text{clean}}^{(l,h)}$ for (a) Gaussian blur and (b) attention-guided occlusion perturbations.}
    \label{fig:drops_combined}
\end{figure}

We visualize this behavior by computing the mean drop in feature count averaged over all layers and heads (Figure~\ref{fig:drops_combined}). For moderate blur, $\Delta C$ fluctuates across relatively low values, rising sharply at the strongest blur setting (Figure \ref{fig:blur_drops}) consistent with the point at which accuracy begins to decline.
For occlusion, $\Delta C$ increases steadily with masking ratio (Figure \ref{fig:occlusion_drops}), with a pronounced rise beyond $60$\% occlusion, coinciding with the point at which classification accuracy also begun to collapse.

Overall, the directional and feature-level analyses show consistent trends under increasing severity. Moderate perturbations preserve both embedding direction and head-wise common-feature counts, while extreme perturbations disrupt both, coinciding with the breakdown in classification accuracy.

\section{Conclusion}
We present a systematic analysis of the representations learned by Masked Autoencoders across both pretraining and fine-tuning. Our analysis shows that pretrained MAE exhibits a clear layer-wise emergence of class-aware structure despite the absence of labels. Visualization of token embeddings shows that class separation strengthens with network depth, and notably, individual patch tokens also become class-discriminative. Subspace analysis reinforces this observation as embeddings from different classes increasingly diverge and occupy distinct subspaces as the network depth increases. The minimum singular value for each class also increases, indicating that even the weakest directions remain informative. We also note that MAE exhibits global attention right from the outset. After fine-tuning, we show that MAE maintains strong classification performance under Gaussian blur and attention-guided occlusion across a wide range of perturbation levels, with a gradual decline at higher severities. This robust behavior is further supported by our results on ImageNet-C dataset. To better understand this robustness, we examine representation robustness from two complementary viewpoints. Under moderate perturbations, directional alignment analysis shows that perturbed embeddings remain closely aligned with their clean counterparts, and head-wise feature-retention analysis reveals that a substantial portion of active features are retained for perturbed inputs. At extreme perturbation levels, both sensitivity indicators degrade sharply, coinciding with a decline in classification accuracy, suggesting a link between the two.

\end{document}